\documentclass[sigconf]{acmart}

\AtBeginDocument{%
  }

\setcopyright{acmlicensed}
\copyrightyear{2026}
\acmYear{2026}
\acmDOI{10.1145/3805622.3810732}
\acmConference[ICMR '26]{ACM International Conference on Multimedia Retrieval}{2026}{Amsterdam,Netherlands}
\acmISBN{979-8-4007-2617-0/26/06}
\usepackage{algorithmic}
\usepackage{graphicx}
\usepackage{booktabs}
\usepackage{hyperref}
\usepackage{textcomp}
\usepackage{amssymb}
\usepackage{bbm}
 
\usepackage{xcolor}
\usepackage{booktabs}   % 提供 \toprule, \midrule, \bottomrule
\usepackage{multirow}   % 提供 \multirow
\usepackage{graphicx}   % 提供 \resizebox
\usepackage{hyperref}
\begin{document}

\title{Relational Retrieval: Leveraging Known-Novel Interactions for Generalized Category Discovery}


% \affiliation{%
\author{Yulin Xu}
\authornote{Both authors contributed equally.}
\affiliation{%
  \institution{University of California, Irvine}
  \city{Irvine}
  \state{CA}
  \country{USA}}
\email{yulinx8@uci.edu}

\author{Chunqi Guo}
\authornotemark[1]
\affiliation{%
  \institution{Sichuan Agricultural University}
  \city{Ya'an}
  \country{China}}
\email{guochunqi02@gmail.com}

\author{Yuanzhen Shuai}
\affiliation{%
  \institution{University College London}
  \city{London}
  \country{UK}}
\email{ucapys0@ucl.ac.uk}

\author{Jianyuan Ni}
\authornote{Corresponding author.}
\affiliation{%
  \institution{Juniata College}
  \city{Huntingdon}
  \country{USA}}
\email{jni100@juniata.edu}

% \affiliation{%

% \affiliation{%

% \affiliation{%

% \affiliation{%

% \affiliation{%

% \affiliation{%

%   \centering  % 添加这一行使图片居中
%   \includegraphics[width=0.8\textwidth]{pipline.png} % 修改宽度为 0.8\textwidth
%   \caption{\textbf{Overview of RPC.} Soft ID/OOD 
%   decomposition via OVA classifiers enables bidirectional        
%   knowledge transfer: embedding fusion for known-class 
%   preservation and relational matching for category        
%   discovery.}
%   \label{fig:frame}

\renewcommand{\shortauthors}{Xu et al.}

\begin{abstract}
In this study, we tackle Generalized Category Discovery (GCD) via a \textbf{Relational Retrieval perspective}, explicitly coupling labeled and unlabeled data through bidirectional knowledge transfer. While existing methods treat these sources separately, missing valuable interaction opportunities, we propose \textbf{\textit{R}}elational \textbf{\textit{P}}attern \textbf{\textit{C}}onsistency (RPC) that enables mutual enhancement. RPC employs One-vs-All classifiers for soft ID/OOD decomposition, then introduces two mechanisms: (i) for known-class preservation, we transfer semantic behavioral alignment; (ii) for category discovery, we leverage the insight that samples from the same category maintain invariant relationships with known-class prototypes, transforming unreliable pseudo-labeling into well-defined relational pattern matching. This bidirectional design allows labeled data to guide unlabeled learning while discovering novel categories through their collective relational signatures. Extensive experiments demonstrate RPC  achieves state-of-the-art performance on both generic and fine-grained benchmarks. 
\end{abstract}


\begin{CCSXML}
<ccs2012>
   <concept>
       <concept_id>10010147.10010178.10010224.10010225.10010231</concept_id>
       <concept_desc>Computing methodologies~Visual content-based indexing and retrieval</concept_desc>
       <concept_significance>300</concept_significance>
       </concept>
 </ccs2012>
\end{CCSXML}

\ccsdesc[300]{Computing methodologies~Visual content-based indexing and retrieval}


\keywords{Generalized category discovery, semi-supervised learning, contrastive learning.}


\maketitle

\section{Introduction}
\label{sec:intro}

% Generalized Category Discovery (GCD)~\cite{vaze2022generalized} addresses a fundamental challenge in open-world recognition: given a
% partially labeled dataset, the goal is to categorize all unlabeled instances, where these instances may belong     
% to either known classes (present in the labeled set) or entirely novel classes never seen during training. This    
% problem setting reflects real-world scenarios where annotation budgets are limited and new categories
% continuously emerge, making it crucial for developing practical vision systems that can both recognize familiar    
% concepts and discover new ones without prior knowledge of the novel class count or their characteristics.

Generalized Category Discovery (GCD)~\cite{vaze2022generalized,he2025category}
extends traditional classification to open-world scenarios by
jointly recognizing known classes and discovering novel
classes in unlabeled data. This problem has been explored
across various tasks, including natural images~\cite{rastegar2023learn,ma2024active,zheng2024prototypical,liu2024novel,wang2025get,zheng2024textual,wu2023metagcd} and specialized applications such as medical imaging and healthcare areas ~\cite{fan2024seeing,zhou2024novel,ni2024survey}, industrial scenarios~\cite{huang2025anomalyncd}, face recognition~\cite{oh2025facegcd,luo2024dig}, information retrieval~\cite{ren2025few}, oracle characters\cite{11482847} etc.

Recent advances in  GCD~\cite{zhao2023gpc,zhao2021novel,wen2023parametric,pu2023dynamic,choi2024contrastive,luo2024contextuality,liu2025debgcd,zuo2025linking} have made significant progress through various strategies, including parametric  classifier learning~\cite{wen2023parametric}, spatial prompt tuning~\cite{wang2024sptnet}, textual information integration~\cite{zheng2024textual,wang2025get}  and soft ID/OOD   
decomposition~\cite{liu2025debgcd}, among others. However, these approaches fundamentally treat labeled and unlabeled data as separate learning streams—labeled data undergoes supervised learning while unlabeled data relies on self-supervised signals. This separation overlooks a critical insight: labeled and unlabeled data can mutually enhance each other through structured interactions, where labeled data provides semantic guidance while unlabeled data offers distributional knowledge. We identify two key limitations of this separation. \textbf{\textit{First,}} unlabeled known-class samples rely solely on self-supervision for representation \cite{chen2020simple}, despite labeled samples of the same classes having already captured highly discriminative features—wasting valuable semantic knowledge. \textbf{\textit{Second,}} category discovery proceeds without considering relational patterns—samples from the same novel category, while semantically distinct from known classes, maintain consistent relationships with known-class  prototypes that serve as powerful grouping signals.

\begin{figure}[t]
\centering
\captionsetup{skip=2pt} % 图和caption之间的距离
\includegraphics[width=1\linewidth]{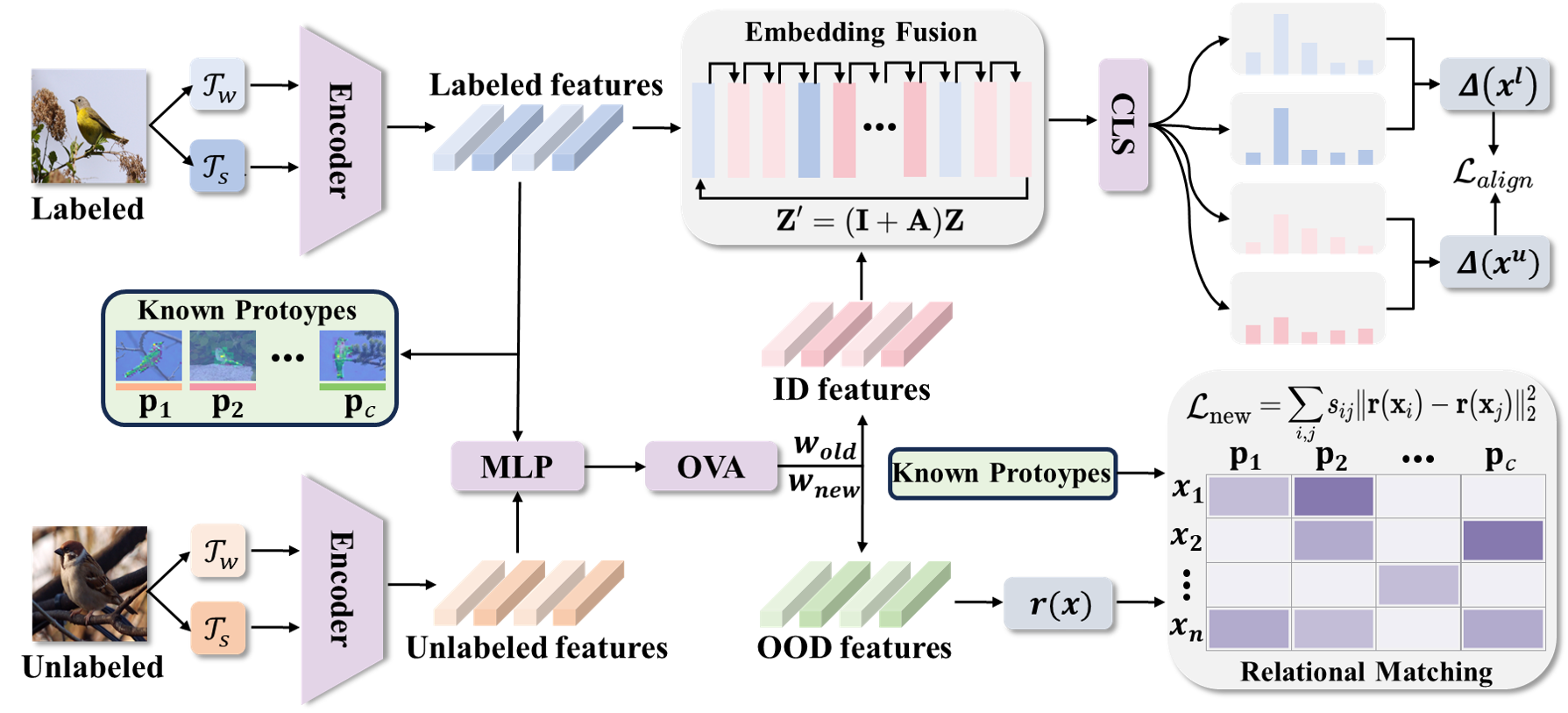}
\caption{\textbf{Overview of RPC.} Soft ID/OOD decomposition via OVA classifiers enables bidirectional knowledge transfer: embedding fusion for known-class  preservation and relational matching for category discovery.}
\label{fig:frame}
\vspace{-0.2cm}
\end{figure}
In this study, we propose Relational Pattern Consistency (RPC), a novel framework that explicitly couples labeled and unlabeled data through bidirectional knowledge transfer as shown in Fig. ~\ref{fig:frame}.  RPC leverages soft ID/OOD decomposition from One-vs-All classifiers  to enable targeted interactions: (i) for known-class preservation, we transfer semantic patterns from labeled to unlabeled data through confidence-weighted embedding fusion and enforce behavioral alignment under augmentation\cite{sohn2020fixmatch}, ensuring unlabeled known-class samples inherit the robust representations from labeled data; (ii) for category discovery, we introduce relational consistency—the principle that samples from the same category maintain invariant relationships with known-class prototype. This transforms discovery from ill-posed clustering into well-defined pattern matching within the known-class reference frame. 

In summary, our contributions are three-fold:
\textcolor{orange}{(i)} We propose RPC, the first GCD framework that couples labeled and unlabeled data through bidirectional knowledge
transfer. Unlike existing methods that maintain separate learning streams, RPC employs soft ID/OOD decomposition to enable targeted interactions. \textcolor{orange}{(ii)} We introduce confidence-weighted embedding fusion with behavioral alignment for known-class preservation, and relational consistency learning that discovers novel categories through their invariant relationships with known-class prototypes, going beyond unreliable pseudo-labels. \textcolor{orange}{(iii)} Extensive experiments demonstrate that RPC achieves state-of-the-art performance on both generic and fine-grained benchmarks, with consistent improvements in known and novel class accuracy validating our bidirectional transfer approach.

\section{Problem Setting}
Let $\mathcal{D}$ denote the dataset, split into a labeled part
$\mathcal{D}_{\mathcal{L}}=\{(\mathbf{x}_i^{l},\mathbf{y}_i^{l})\}\subset\mathcal{X}\times\mathcal{Y}_{l}$
and an unlabeled part
$\mathcal{D}_{\mathcal{U}}=\{\mathbf{x}_i^{u}\}\subset\mathcal{X}$, where each
$\mathbf{x}_i^{u}$ is associated with a hidden ground-truth label
$\hat{\mathbf{y}}_i^{u}\in\mathcal{Y}_{u}$ used only for evaluation.
The unlabeled pool may contain categories absent from the labeled subset, i.e.,
$\mathcal{Y}_{l}\subset\mathcal{Y}_{u}$. The learning objective is to assign
every instance in $\mathcal{D}_{\mathcal{U}}$ to a semantic category.
Following~\cite{wen2023parametric}, we assume the total number of categories in
the unlabeled set is known a priori, and write $K=|\mathcal{Y}_{u}|$.

\section{Method}

	\subsection{Baseline for GCD}
In this study, our model builds on SimGCD~\cite{wen2023parametric} for representation learning and parametric classification,  while incorporating DebGCD's~\cite{liu2025debgcd} One-vs-All (OVA) decomposition to identify whether unlabeled samples belong to known or novel classes. This combination enables effective feature
learning through SimGCD while OVA decomposition provides reliable soft weights for our bidirectional knowledge transfer.
	
	\subsubsection{Representation Learning}
	\label{sec:simgcd}
	
	The objective of representation learning is to obtain discriminative features, which allows the classifier effectively classify all the categories.
	To do so, a pre-trained feature extractor $f(\cdot)$ is fine-tuned with two contrastive losses. Given two random augmentations $\hat{\mathbf{x}}_i$ and $\tilde{\mathbf{x}}_i$ of an image $\mathbf{x}_i$ in a training mini-batch $\mathcal{B}$ that contains both labeled and unlabeled data, the self-supervised contrastive loss is formulated as: 
	{\small
    \begin{equation}
		\mathcal{L}_{\text{rep}}^{u}=\frac{1}{|\mathcal{B}|} \sum_{\mathbf{x}_i\in \mathcal{B}} - \log \frac{\exp (\hat{\mathbf{z}}_i^\top \tilde{\mathbf{z}}_i / \tau_u)}{\sum_{\mathbf{x}_j \in B} \exp (\hat{\mathbf{z}}_j^\top \tilde{\mathbf{z}}_i / \tau_u)},
	\end{equation}
    }
	where $\mathbf{z}_i=g(f(\mathbf{x}_i))$ is the projected feature for contrastive learning, $g(.)$ is the projection head, and $\tau_u$ is a temperature value.
	% We also makes use of a \emph{supervised} contrastive loss to learn a representation from the labeled data:
	A supervised contrastive loss is similarly defined by:
	{\small
    \begin{equation}
		\mathcal{L}_{\text{rep}}^{s} = \frac{1}{|\mathcal{B}^l|}\sum_{\mathbf{x}_i\in \mathcal{B}^l}\frac{1}{|\mathcal{N}_i|}\sum_{p\in\mathcal{N}_i}-\log\frac{\exp (\hat{\mathbf{z}}_i^\top \tilde{\mathbf{z}}_p / \tau_s)}{\sum_{n\neq i}\exp (\hat{\mathbf{z}}_i^\top \tilde{\mathbf{z}}_n / \tau_s)},
	\end{equation}
    }
	where $\mathcal{N}_i$ is the indexes of images in the mini-batch that have the same label to $\mathbf{x}_i$, $B^l$ is the mini-batch of labeled training data and $\tau_s$ is a temperature value. The two losses are combined to learn the representation: $\mathcal{L}_{\text{rep}}=(1-\lambda) \mathcal{L}_{\text{rep}}^{u} + \lambda \mathcal{L}_{\text{rep}}^{s}$, where $\lambda$ is a hyperparameter.
	
	\subsubsection{Classifier Learning}
	% Classifier learning aims to learn a classifier that can assign labels to all unlabeled data.
	% In addition to the representation learning, we also follow SimGCD~\cite{wen2023parametric} to set up the classifier learning process.
	Following SimGCD~\cite{wen2023parametric}, we employ a    
  parametric classifier with learnable prototypes
  $\mathcal{T}=\{\mathbf{t}_1, \mathbf{t}_2, \dots,
  \mathbf{t}_K\}$, where $K=|\mathcal{Y}_u|$ is assumed
  known a priori. During training, the soft label
  $\hat{\mathbf{p}}_i^k$ for each augmented view
  $\mathbf{x}_i$ is computed via softmax over cosine
  similarities between features and prototypes:
  {\footnotesize
  \begin{equation}
      \hat{\mathbf{p}}_i^k =
  \frac{\exp\left(({\hat{\mathbf{h}}_i} /
  {\|\hat{\mathbf{h}}_i\|_2})^\top ({\mathbf{t}_k}/{\|\mathbf{t}_k\|_2})/\tau_s\right)}{\sum_{j}\exp{\left((\hat{\mathbf{h}}_i/\|\hat{\mathbf{h}}_i\|_2)^\top
  (\mathbf{t}_j/\|\mathbf{t}_j\|_2)/\tau_s\right)}},
  \end{equation}
  }

 where $\hat{\mathbf{h}}_i=f(\hat{\mathbf{x}}_i)$.
  Similarly computing $\tilde{\mathbf{p}}_i$ for
  $\tilde{\mathbf{x}}_i$, the classifier losses become:
{\small
\begin{equation}
\label{Cls loss}
\mathcal{L}_{\text{cls}}^s = \frac{1}{|\mathcal{B}^l|}\sum_{\mathbf{x}_i \in \mathcal{B}^l} \ell_{CE}(\mathbf{y}({x_i}), \hat{\mathbf{p}}_i), \,
\mathcal{L}_{\text{cls}}^u = \frac{1}{|\mathcal{B}|}\sum_{\mathbf{x}_i \in \mathcal{B}} \ell_{CE}(\tilde{\mathbf{p}}_i, \hat{\mathbf{p}}_i),
\end{equation}
}
	where $\mathbf{y}_i$ is the ground truth label for the labeled data point $\mathbf{x}_i$, $\mathcal{L}_{\text{ce}}$ is the cross-entropy loss, and $H(\overline{\mathbf{p}})=-\sum \overline{\mathbf{p}} \log \overline{\mathbf{p}}$ regularizes the mean prediction $\overline{\mathbf{p}}=\frac{1}{2|B|}\sum_{\mathbf{x}_i\in B}(\hat{\mathbf{p}}_i + \tilde{\mathbf{p}}_i)$ in a mini-batch.
	The final classifier loss is given by $\mathcal{L}_{\text{cls}}=(1-\lambda) \mathcal{L}_{\text{cls}}^u+\lambda \mathcal{L}_{\text{cls}}^l- \epsilon H(\overline{\mathbf{p}})$. 
	 
	Combining the losses of the representation learning and the classifier learning, the overall loss of the baseline SimGCD is formulated by:
		$\mathcal{L}_{\text{baseline}} = \mathcal{L}_{\text{rep}} + \mathcal{L}_{\text{cls}}.$

\subsubsection{One-vs-All Decomposition}

\begin{table*}[t]
\captionsetup{skip=2pt}
  \caption{Comparison with state-of-the-art methods on generic and fine-grained benchmarks. All values are percentages.}
  \centering
  \footnotesize
  \tabcolsep=0.15cm
  % 建议直接使用 1.0\textwidth，适配跨栏表格
  \resizebox{1.0\textwidth}{!}{
  \begin{tabular}{l ccc ccc ccc ccc ccc ccc}
  \toprule
      \multirow{2}{*}{Methods} & 
      \multicolumn{3}{c}{CUB} & \multicolumn{3}{c}{Stanford Cars} & 
      \multicolumn{3}{c}{FGVC-Aircraft} & 
      \multicolumn{3}{c}{CIFAR10} & \multicolumn{3}{c}{CIFAR100} & \multicolumn{3}{c}{ImageNet-100}      
  \\
  \cmidrule(lr){2-4} \cmidrule(lr){5-7} \cmidrule(lr){8-10} 
  \cmidrule(lr){11-13} \cmidrule(lr){14-16} \cmidrule(lr){17-19}
      & All & Old & New  & All  & Old & New & All & Old & New & All & Old & New  & All & Old & New & All & Old & New \\
  \midrule % 建议使用 \midrule 替代 \hline，符合 booktabs 风格
      k-means~\cite{macqueen1967some_kmeans} & 34.3 & 38.9 & 32.1 & 12.8 & 10.6 & 13.8 & 16.0 & 14.4 & 16.8 & 83.6 & 85.7 & 82.5 & 52.0 & 52.2 & 50.8 & 72.7 & 75.5 & 71.3 \\
      RS+~\cite{han2021autonovel} & 33.3 & 51.6 & 24.2 & 28.3 & 61.8 & 12.1 & 26.9 & 36.4 & 22.2 & 46.8 & 19.2 & 60.5 & 58.2 & 77.6 & 19.3 & 37.1 & 61.6 & 24.8 \\
      UNO+~\cite{fini2021unified} & 35.1 & 49.0 & 28.1 & 35.5 & 70.5 & 18.6 & 40.3 & 56.4 & 32.2 & 68.6 & \textbf{98.3} & 53.8 & 69.5 & 80.6 & 47.2 & 70.3 & \textbf{95.0} & 57.9 \\
      \midrule
      GCD~\cite{vaze2022generalized} & 51.3 & 56.6 & 48.7 & 39.0 & 57.6 & 29.9 & 45.0 & 41.1 & 46.9 & 91.5 & \underline{97.9} & 88.2 & 73.0 & 76.2 & 66.5 & 74.1 & 89.8 & 66.3 \\
      PromptCAL~\cite{zhang2023promptcal} & 62.9 & 64.4 & 62.1 & 50.2 & 70.1 & 40.6 & 52.2 & 52.2 & 52.3 & \textbf{97.9} & 96.6 & 98.5 & 81.2 & 84.2 & 75.3 & 83.1 & 92.7 & 78.3 \\
      DCCL~\cite{pu2023dynamic} & 63.5 & 60.8 & \underline{64.9} & 43.1 & 55.7 & 36.2 & - & - & - & 96.3 & 96.5 & 96.9 & 75.3 & 76.8 & 70.2 & 80.5 & 90.5 & 76.2 \\
      SimGCD~\cite{wen2023parametric} & 60.3 & 65.6 & 57.7 & 53.8 & 71.9 & 45.0 & 54.2 & 59.1 & 51.8 & 97.1 & 95.1 & 98.1 & 80.1 & 81.2 & 77.8 & 83.0 & 93.1 & 77.9 \\
      SPTNet~\cite{wang2024sptnet} & 65.8 & 68.8 & \textbf{65.1} & 59.0 & 79.2 & 49.3 & 59.3 & 61.8 & 58.1 & 97.3 & 95.0 & \underline{98.6} & 81.3 & 84.3 & 75.6 & 85.4 & 93.2 & 81.4 \\
      ProtoGCD~\cite{ma2025protogcd} & 63.2 & 68.5 & 60.5 & 53.8 & 73.7 & 44.2 & 56.8 & 62.5 & 53.9 & 
97.3 & 95.3 & 98.2 & 81.9 & 82.9 & 80.0 & 84.0 & 92.2 & 79.9 \\
      DebGCD~\cite{liu2025debgcd} & \underline{66.3} & \underline{71.8} & 63.5 & \underline{65.3} & \textbf{81.6} & \underline{57.4} & \underline{61.7} & \underline{63.9} & \underline{60.6} & 97.2 & 94.8 & 98.4 & \underline{83.0} & \underline{84.6} & \underline{79.9} & \underline{85.9} & \underline{94.3} & \underline{81.6} \\
    
      RPC (Ours) & \textbf{67.1} & \textbf{73.2} & 64.8 & \textbf{65.5} & \underline{81.0} & \textbf{57.8} & \textbf{62.1} & \textbf{65.5} & \textbf{61.0} & \underline{97.6} & 95.2 & \textbf{98.8} & \textbf{83.8} & \textbf{85.1}  & \textbf{81.0} & \textbf{86.2} & 93.8 & \textbf{82.1} \\
  \bottomrule
  \end{tabular}
  }
  \vspace{-0.5 cm}
  \label{table:main_results}
\end{table*}

 Following DebGCD~\cite{liu2025debgcd}, we employ One-vs-All (OVA) classifiers for soft ID/OOD
  decomposition. For each known class $c \in \mathcal{Y}_l$, we pre-train a binary classifier $h_c:
  \mathbb{R}^d \rightarrow [0,1]$ to distinguish class $c$ from all others. These classifiers are used to     
  compute ID/OOD scores $s_{ID}(\mathbf{x}^u)$ for unlabeled samples, yielding soft weights:
 {\small
 \begin{equation}
  w_\text{old}(\mathbf{x}^u) = s_{ID}(g(f(\mathbf{x}^u))), \quad w_\text{new}(\mathbf{x}^u) = 1 -
  s_{ID}(g(f(\mathbf{x}^u))),
  \end{equation}
 }
  where $w_\text{old}$ indicates the likelihood of belonging to known classes and $w_\text{new}$ to novel     
  classes. These weights guide our bidirectional knowledge transfer: samples with high $w_\text{old}$
  participate in semantic pattern transfer from labeled data, while samples with high $w_\text{new}$
  contribute to relational consistency learning for novel discovery.

\subsection{Relational Pattern Consistency (RPC)}

Unlike prior GCD methods that treat labeled and unlabeled     
data as two disjoint domains, RPC explicitly
\emph{couples} them through bidirectional knowledge
transfer, as shown in Fig. ~\ref{fig:frame}. Our approach rests on two key insights:
\textbf{(i)} labeled data contains rich semantic structure    
that can guide unlabeled known-class learning beyond
unreliable pseudo-labels, and \textbf{(ii)} samples from      
the same category exhibit consistent behavioral signals       
when interacting with known-class prototypes, and
critically, labeled data provides the reference framework     
to capture and interpret these behavioral patterns for        
category discovery.

\subsubsection{Known-Class Preservation via Semantic Transfer} 

 For known-class preservation, we transfer robust        
  semantic patterns from labeled to unlabeled data        
  through behavioral consistency under augmentation.      
We construct training batches by selecting ID-likely unlabeled samples based on OVA scores. For each batch of $B$ labeled samples, we first sample $\mu$ unlabeled samples and compute $\mu_{\text{ID}} = \lfloor \mu \cdot \rho_{\text{ID}} \rfloor$, where $\rho_{\text{ID}}$ is the estimated proportion of ID-class samples in the unlabeled set. We then select the top-$\mu_{\text{ID}}$ samples with highest $w_{\text{known}}$ scores from these $\mu$ samples, resulting in the pattern: $[\mathbf{x}_1^l, \mathbf{x}_{1,1}^u, ..., \mathbf{x}_{1,\mu_{\text{ID}}}^u, \mathbf{x}_2^l, \mathbf{x}_{2,1}^u, ..., \mathbf{x}_{2,\mu_{\text{ID}}}^u, ...]$, where $\mathbf{x}_{i,j}^u$ denotes the $j$-th ID-class unlabeled sample paired with the $i$-th labeled sample, all sharing identical augmentations.

To enable semantic transfer from labeled to unlabeled samples, we apply confidence-weighted embedding\cite{huang2024interlude}. Given batch embeddings $\mathbf{Z} \in \mathbb{R}^{Q \times d}$ where $Q = B(1+\mu_{\text{ID}})$, we compute:

  \begin{equation}
  \mathbf{Z}' = (\mathbf{I} + \mathbf{A})\mathbf{Z}       
  \end{equation}

Following the circular shift design, the fusion matrix is:
\begin{equation}
    \mathbf{A}_{ij} = \begin{cases}
\alpha \cdot w_{\text{old}}(\mathbf{x}_i) & \text{if } j = (i-1) \mod Q \\
-\alpha \cdot w_{\text{old}}(\mathbf{x}_i) & \text{if } i = j \\
0 & \text{otherwise}
\end{cases}
\end{equation}
This design ensures that each unlabeled sample $\mathbf{x}_{k,m}^u$ (the $m$-th unlabeled sample paired with the $k$-th labeled sample) receives information from its preceding sample in the batch, which is either the paired labeled sample (when $m=1$) or another unlabeled sample. The weighting by $w_{\text{old}}$ ensures that only high-confidence ID samples are significantly affected by the fusion.

With these fused embeddings $\mathbf{Z}'$, we enforce behavioral consistency between each labeled sample and its paired unlabeled samples:
 \begin{equation}
 % \small
  \mathcal{L}_{align} = \frac{1}{|\mathcal{B}^l|} \sum_{i=1}^{|\mathcal{B}^l|}        
  \left\| \Delta(\mathbf{x}_i^l) -
  \frac{\sum_{j=1}^{\mu} w_{old}(\mathbf{x}_{ij}^u)     
  \cdot \Delta(\mathbf{x}_{ij}^u)}{\sum_{j=1}^{\mu}       
  w_{old}(\mathbf{x}_{ij}^u)} \right\|_2^2,
  \end{equation}
where $\Delta(\mathbf{x}) = g(f(\mathcal{T}_w(\mathbf{x}))) - g(f(\mathcal{T}_s(\mathbf{x})))$ captures augmentation-induced behavioral changes, with $\mathcal{T}_w$ and $\mathcal{T}_s$ denoting weak and strong augmentations respectively. For labeled samples, $\Delta(\mathbf{x}^l)$ is computed on original features, while for unlabeled samples, $\Delta'(\mathbf{x}^u)$ is computed on fused features from $\mathbf{Z}'$.

\subsubsection{Category Discovery via Relational Matching}
For category discovery in this step, we leverage a key insight: while novel classes are semantically distinct from known classes, samples from the same novel category usually share consistent patterns of similarity and dissimilarity with the known-class reference system. Specifically, if two samples belong to the same novel class, they should exhibit similar distances to each known-class prototype—being similarly far from some known classes and similarly close to others. This shared relational pattern serves as an implicit signature of category membership.

To operationalize this, we encode each sample's relational signature as its similarity pattern with all known-class prototypes:
  {\small
  \begin{equation}
      \mathbf{r}(\mathbf{x}) = \left[\frac{f(\mathbf{x})        
  \cdot \mathbf{p}_1}{\|f(\mathbf{x})\|_2
  \|\mathbf{p}_1\|_2}, ..., \frac{f(\mathbf{x}) \cdot
  \mathbf{p}_{C_L}}{\|f(\mathbf{x})\|_2
  \|\mathbf{p}_{C_L}\|_2}\right] \in \mathbb{R}^{C_L},
  \end{equation}
  }
where $\{\mathbf{p}_c\}_{c=1}^{C_L}$ are known-class prototypes. This transforms each sample into coordinates within the known-class reference frame.
We enforce relational consistency among feature-similar samples:
\begin{equation}
    \mathcal{L}_{\text{new}} = \sum_{i,j \in \mathcal{B}^u} w_{\text{new}}(\mathbf{x}_i^u) \cdot w_{\text{new}}(\mathbf{x}_j^u) \cdot s_{ij} \cdot \|\mathbf{r}(\mathbf{x}_i^u) - \mathbf{r}(\mathbf{x}_j^u)\|_2^2,
\end{equation}
where $s_{ij} = \exp(\langle f(\mathbf{x}_i^u), f(\mathbf{x}_j^u) \rangle / (\|f(\mathbf{x}_i^u)\|_2 \|f(\mathbf{x}_j^u)\|_2 \cdot \tau_u))$ weights pairs by feature similarity. This transforms novel discovery from ill-posed clustering into well-defined pattern matching—samples from the same novel category share consistent relationships with all known classes.

\subsection{Training Objective}

The complete RPC objective integrates baseline learning with our bidirectional consistency mechanisms:

\begin{equation}
\mathcal{L}_{total} = \mathcal{L}_{baseline} +  \lambda_1 \mathcal{L}_{align} + \lambda_2 \mathcal{L}_{new}
\end{equation}
 where $\lambda_1$ and $\lambda_2$ balance known-class preservation and novel class  relational consistency respectively. 

\section{Experiments}

\noindent
\textbf{Datasets}. We conduct experiments on six benchmarks covering diverse scenarios. Three fine-grained datasets from the Semantic Shift Benchmark (SSB)— CUB~\cite{welinder2010}, Stanford Cars~\cite{krause2013}, and FGVC-Aircraft~\cite{maji2013}—are used to evaluate
performance on specialized domains. Two general benchmarks, CIFAR10/100~\cite{krizhevsky2009} and ImageNet-100~\cite{tian2020a}, assess broader object recognition capabilities. Consistent with the GCD setting~\cite{vaze2022generalized}, we split categories into known ($\mathcal{Y}_l$) and novel classes, where 50\% of known-class images form the labeled data $\mathcal{D}_\mathcal{L}$ and all other images constitute the unlabeled pool $\mathcal{D}_\mathcal{U}$.

\noindent \textbf{Implementation details}.
Following SimGCD~\cite{wen2023parametric}, we adopt a ViT-B/16 backbone pre-trained with DINO~\cite{caron2021}, consistent with the set up in~\cite{vaze2022generalized}. Image features, $f(\cdot)$, are extracted from the 768-dimensional [CLS] token. We fine-tune only the final block with a batch size of 128 over 200 epochs, starting with a learning rate of 0.1 and applying cosine learning rate decay. Following ~\cite{wen2023parametric}, the balance factor $\lambda$ is 0.35, and the temperature parameters $\tau_s$ and $\tau_u$ are set to 0.1 and 0.07, respectively.  In this study, we empirically set $\lambda_1 = 0.5$, $\lambda_2 = 0.3$, and the
embedding fusion weight $\alpha = 0.3$ based on hyperparameter analysis.

\noindent\textbf{Compared methods.} 
The compared methods include three competing novel category discovery (NCD) methods 
RS+~\cite{han2021autonovel}, 
UNO+~\cite{fini2021unified}, 
and k-means~\cite{macqueen1967some_kmeans} 
with DINO~\cite{caron2021} features,  and several state-of-the-art generalized category discovery (GCD) methods, including 
GCD~\cite{vaze2022generalized}, 
PromptCAL~\cite{zhang2023promptcal}, 
DCCL~\cite{pu2023dynamic}, 
SimGCD~\cite{wen2023parametric},  SPTNet~\cite{wang2024sptnet}, ProtoGCD~\cite{ma2025protogcd} and DebGCD~\cite{liu2025debgcd}. Importantly, Our method builds on SimGCD's~\cite{wen2023parametric}        
framework and DebGCD's~\cite{liu2025debgcd} OVA
decomposition, adding bidirectional knowledge transfer.  

\begin{table}[t]
\captionsetup{skip=2pt}
  \centering
  \small
  \caption{Ablation study on CUB and ImageNet-100.}
  \label{tab:ablation}
  \begin{tabular}{l ccc ccc}
    \toprule
    & \multicolumn{3}{c}{CUB} & \multicolumn{3}{c}{ImageNet-100} \\
    \cmidrule(lr){2-4} \cmidrule(lr){5-7}
    Methods & All & Old & New & All & Old & New \\
    \midrule
    w/o Embedding Fusion           & 64.9 & 70.1 & 62.4 & 84.0 & 90.6 & 81.2 \\
    w/o $\mathcal{L}_{\text{align}}$    & 65.6 & 71.0 & 63.6 & 84.6 & 91.4 & 81.5 \\
    w/o $\mathcal{L}_{\text{discover}}$ & 65.2 & 73.0 & 61.5 & 84.3 & 92.6 & 79.0 \\
    \midrule
    RPC (Ours)                     & \textbf{67.1} & \textbf{73.2} & \textbf{64.8} & \textbf{86.2} & \textbf{93.8} & \textbf{82.1} \\
    \bottomrule
  \end{tabular}
\vspace{-0.4 cm}
\end{table}

\begin{table} [t]
\captionsetup{skip=1pt}
\centering
\label{tab:overhead}
\small
\caption{Computational overhead comparison on CUB.} % 建议加上数据集名称更严谨
\begin{tabular}{lcc}
\toprule
Method & Params (M) & FLOPs (G) \\
\midrule
SimGCD & 92.25 & 17.60 \\
RPC (Ours) & \textbf{93.54} {\small
(+1.40\%)} & \textbf{17.92} {\small (+1.82\%)} \\
\bottomrule
\end{tabular}
\vspace{-0.2 cm}
\end{table}
 
\subsection{Results and Discussion}
	
% \noindent\textbf{Compared methods.}
% We evaluate against novel class discovery baselines
% (RS+~\cite{han2021autonovel}, UNO+~\cite{fini2021unified},    
% k-means~\cite{macqueen1967some_kmeans}) and six
% state-of-the-art GCD methods~\cite{vaze2022generalized,zhang2023promptcal,pu2023dynamic,wen2023parametric,wang2024sptnet,liu2025debgcd}.     
% Our method builds on SimGCD's~\cite{wen2023parametric}        
% framework and DebGCD's~\cite{liu2025debgcd} OVA
% decomposition, adding bidirectional knowledge transfer.  

\noindent\textbf{Results analysis.}
  Tab. \ref{table:main_results} shows RPC achieves state-of-the-art performance across most benchmarks. Three key findings emerge:  (1) Old ACC improves significantly (\textbf{73.2\%} on CUB compared to DebGCD's \textbf{71.8\%}), validating our semantic transfer mechanism. (2) New ACC gain \textbf{1.3\%} on CUB and\textbf{ 0.5\%}      
  on ImageNet-100, showing relational consistency's
  effectiveness. (3) Fine-grained datasets yield larger
  overall improvements (\textbf{0.8\%} on CUB, \textbf{0.4\%} on Aircraft) than generic ones (\textbf{0.3\%} on ImageNet-100), indicating relational patterns better capture subtle distinctions.

\noindent\textbf{Ablation Study}
Tab.~\ref{tab:ablation} validates each component's contribution. Removing embedding fusion or $\mathcal{L}_{\text{align}}$ reduces known-class accuracy by \textbf{3.1\%} and \textbf{3.2\%} on CUB respectively, confirming their role in semantic transfer. Without $\mathcal{L}_{\text{new}}$, novel-class accuracy drops \textbf{3.3\%} on CUB, demonstrating its effectiveness for category discovery. All components prove essential for optimal performance.

\begin{figure}
\captionsetup{skip=2pt}
    \centering
    \includegraphics[width=1\linewidth]{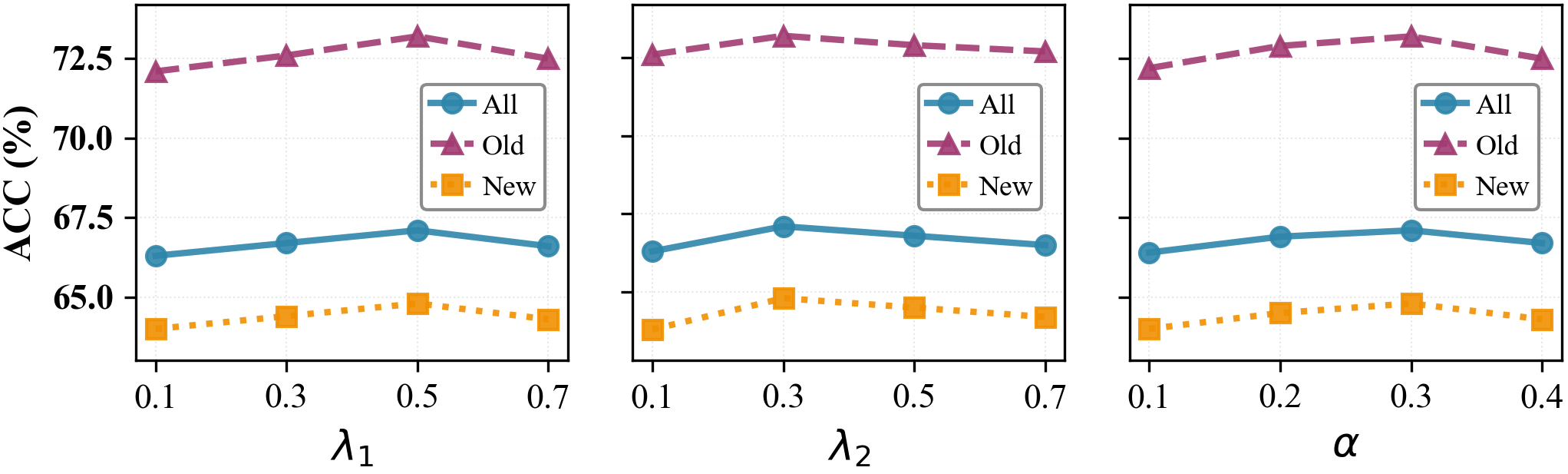}
    \caption{Impact of hyperparameters $\lambda_1$, $\lambda_2$, and $\alpha$ on CUB.}
    \label{fig:hyper}
    \vspace{-0.5 cm}
\end{figure}

\noindent\textbf{Hyperparameter Analysis.}  Fig. \ref{fig:hyper} shows RPC's robustness on CUB. Optimal  values ($\lambda_1$=0.5, $\lambda_2$=0.3, $\alpha$=0.3) balance known-class preservation, category discovery, and feature fusion. Performance degrades gracefully beyond these  points—excessive $\lambda_1$ over-constrains adaptation, high $\lambda_2$ causes unstable  classification, and large $\alpha$ leads to over-fusion. Overall variation remains within \textbf{1\%},  validating our method's stability.

\noindent\textbf{Computational Efficiency Analysis.} To evaluate the practical applicability of the proposed RPC, we compare its computational overhead with the baseline SimGCD. As summarized in Table 3, RPC introduces a minimal increase in model complexity. Specifically, our method only adds 1.29M parameters (a 1.40\% increase) and 0.32G FLOPs (a 1.82\% increase) over the baseline.

\begin{figure}
    \centering
    \includegraphics[width=1\linewidth]{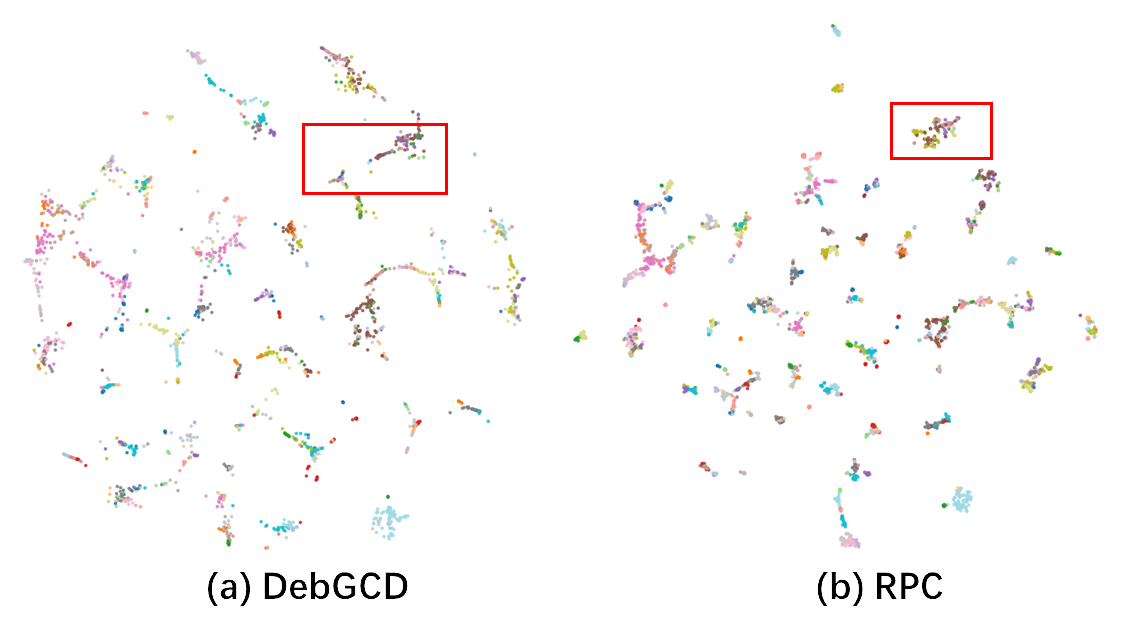}
    \caption{t-SNE visualization of feature embeddings on cub dataset. (a) DebGCD  and (b) our RPC. The red boxes highlight specific regions where our method achieves significantly tighter intra-class grouping and clearer inter-class boundaries. Compared to DebGCD, RPC effectively eliminates the semantic entanglement between neighboring categories.}
    \label{fig:tsne}
     \vspace{-0.5 cm}
\end{figure}

\noindent\textbf{Feature Visualization.} To qualitatively evaluate the quality of the learned representations, we visualize the feature embeddings using t-SNE. As illustrated in Fig. \ref{fig:tsne}, RPC produces a more structured and discriminative latent space compared to DebGCD. Specifically, within the areas highlighted by red boxes, it is observable that RPC effectively compresses the intra-class variance while expanding the inter-class margins. This superior separability suggests that our Relational Pattern Consistency (RPC) framework better preserves the underlying semantic structure, which is crucial for identifying novel categories in a shared embedding space.

% To verify the effectiveness of our model in learning cross-instance features, we present in Tab.~\ref{tab:ablation}. We dissect RPC along four axes: (1) replacing our prototype–based relational matching with pairwise feature consistency yields clear drops, especially on novel classes; (2) using learned prototypes as anchors is more robust than using individual samples; (3) confidence–weighted (soft) prototypes outperform fixed mean (hard) prototypes; and (4) asymmetric augmentation strengths across the two branches (labeled/unlabeled) outperform identical augmentations. Removing embedding fusion or alignment further degrades known-class accuracy, while the complete RPC delivers the best balance on both old and new categories.

\section{Conclusion}

In this study, we introduced bidirectional knowledge transfer for Generalized Category Discovery (GCD) tasks. We proposed Relational Pattern Consistency (RPC), which transfers semantic patterns from labeled to unlabeled data through embedding fusion, while discovering novel categories via relational consistency with known-class prototypes. This transforms GCD from parallel learning into mutual enhancement between labeled and unlabeled data. Extensive experiments on six benchmarks demonstrate the competitiveness of the proposed RPC methods.
 
%\clearpage
\bibliographystyle{ACM-Reference-Format}
\bibliography{sample-base}

\appendix

\end{document}